\documentclass[10pt,twocolumn,letterpaper]{article}

\usepackage{cvpr}      

\usepackage[dvipsnames]{xcolor}

\definecolor{cvprblue}{rgb}{0.21,0.49,0.74}
\usepackage[pagebackref,breaklinks,colorlinks,citecolor=cvprblue]{hyperref}

\usepackage{cuted}
\usepackage{capt-of}

\def\paperID{8593} 
\def\confName{CVPR}
\def\confYear{2024}

\title{Bootstrap Fine-Grained Vision-Language Alignment for Unified Zero-Shot Anomaly Localization}

\author{Hanqiu Deng, Zhaoxiang Zhang, Jinan Bao, Xingyu Li\\
University of Alberta
}

\begin{document}

\maketitle

\begin{strip}
\centering
\includegraphics[width=\linewidth]{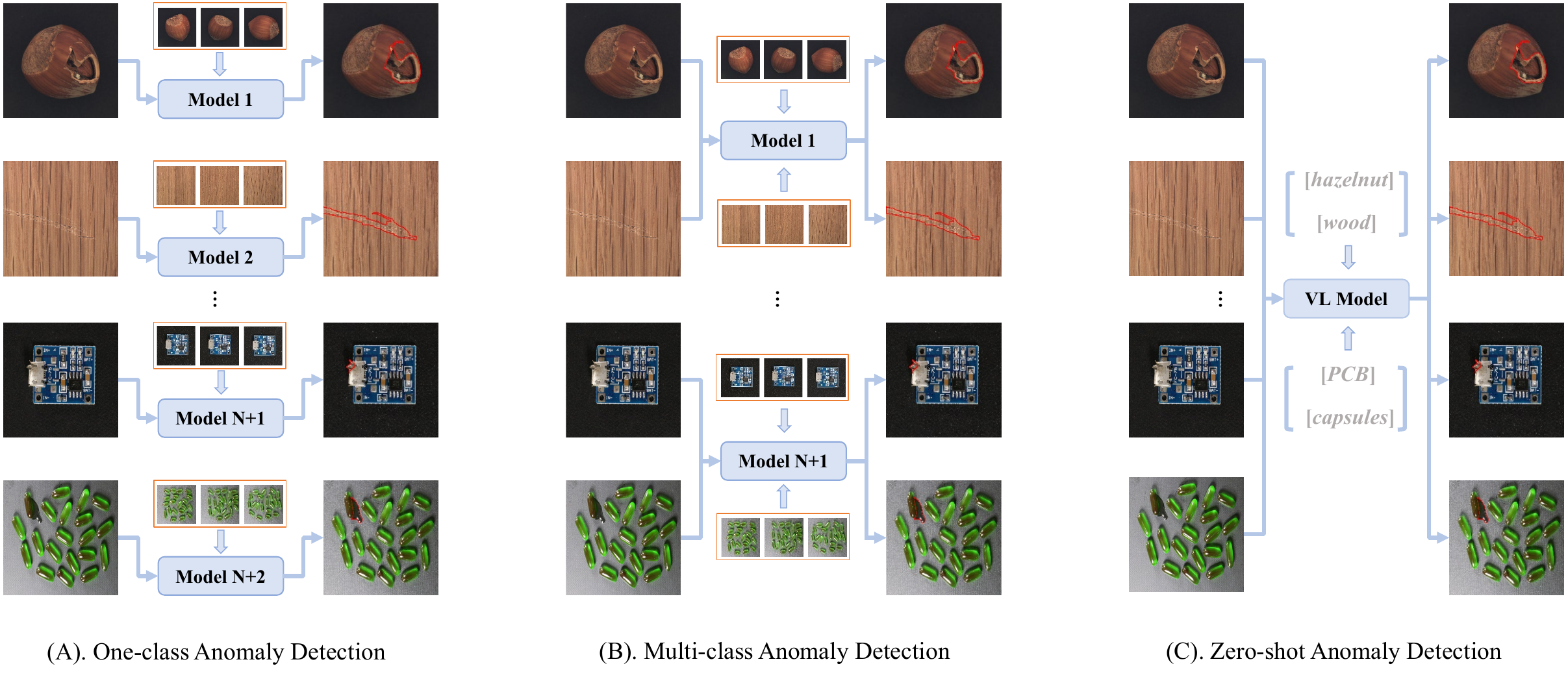} 
\captionof{figure}{A review of present visual anomaly detection tasks. Fig. (A) shows that anomaly detection is proposed as a one-class classification task \cite{mvtec,visa}, where a model is trained for each class in the dataset. As shown in Fig. (B), multi-class anomaly detection is proposed to improve the efficiency of model usage by training a model on normal images from multiple categories \cite{uniad}. Fig. (C) shows a new challenging task, zero-sample anomaly detection, that allows the model to localize anomalous regions without touching any normal samples in any category. In this study, we focus on tackling this task with vision-language (VL) models \cite{clip} that exhibit open-world intelligibility.}
\label{anovl_teaser}
\end{strip}

\begin{abstract}
Contrastive Language-Image Pre-training (CLIP) models have shown promising performance on zero-shot visual recognition tasks by learning visual representations under natural language supervision. Recent studies attempt the use of CLIP to tackle zero-shot anomaly detection by matching images with normal and abnormal state prompts. However, since CLIP focuses on building correspondence between paired text prompts and global image-level representations, the lack of fine-grained patch-level vision to text alignment limits its capability on precise visual anomaly localization. In this work, we propose AnoCLIP for zero-shot anomaly localization. In the visual encoder, we introduce a training-free value-wise attention mechanism to extract intrinsic local tokens of CLIP for patch-level local description. From the perspective of text supervision, we particularly design a unified domain-aware contrastive state prompting template for fine-grained vision-language matching. On top of the proposed AnoCLIP, we further introduce a test-time adaptation (TTA) mechanism to refine visual anomaly localization results, where we optimize a lightweight adapter in the visual encoder using AnoCLIP's pseudo-labels and noise-corrupted tokens. With both AnoCLIP and TTA, we significantly exploit the potential of CLIP for zero-shot anomaly localization and demonstrate the effectiveness of AnoCLIP on various datasets. 
\end{abstract}

\section{Introduction}

Visual anomaly detection and localization is a pivotal topic of computer vision, evident in its diverse applications across industrial visual inspection \cite{mvtec, visa}, video surveillance \cite{video}, and medical diagnosis \cite{medical}. This intricate and fine-grained task involves the detection and localization of atypical patterns, anomalies, or deviations within visual data. Its complexity nature mainly stems from the inherent diversity of anomalies. For instance, many anomalies manifest as subtle deviations in texture, color, shape, or motion, blending into the surrounding normal samples \cite{mvtec, visa}.


Prior arts in anomaly detection and localization focus on training a dedicated model for each category of normal images \cite{simple, core, rd, us, draem}. However, training category-specific models is undoubtedly costly in real-world scenarios. 
Recently, impressive progress has been made towards unified models for multi-class anomaly detection, aiming to establish a single model capable of identifying multi-class anomalies \cite{uniad,regad}. Nonetheless, these advancements are still limited to modeling normal distributions on known categories and fail in detecting anomalies in unknown or open-set scenarios. In addition both ways have limited the scalability of the model, and in practice it is still necessary to retrain the model for incoming categories. In this study, we tackle an open-world anomaly detection problem, which is to identify and locate anomalies on unknown classes of samples in a zero-shot manner. As shown in Fig. \ref{anovl_teaser}, we show one-class and multi-class anomaly detection methods that rely on normal image training, as well as our proposed unified zero-shot anomaly detection paradigm.

Recent CLIP models have shown promising properties for detecting unknown objects \cite{vild, kuo2022f} as well as out-of-distribution data \cite{ood, cohen2022out}. By learning on millions of image-text pairs, CLIP has excellent zero-shot image classification ability. In this study, we aim to achieve zero-shot anomaly localization with CLIP. To this end, two challenges in CLIP have to be well addressed. 
First, anomaly detection is a fine-grained task where the discriminative details are too subtle to be well represented by CLIP. Second, CLIP presents limited local descriptors due to its training on paring image-wise global features and text embeddings. This becomes an obstacle to precisely locating anomalous regions. 

\begin{figure*}[!ht]
\centering
\includegraphics[width=\textwidth]{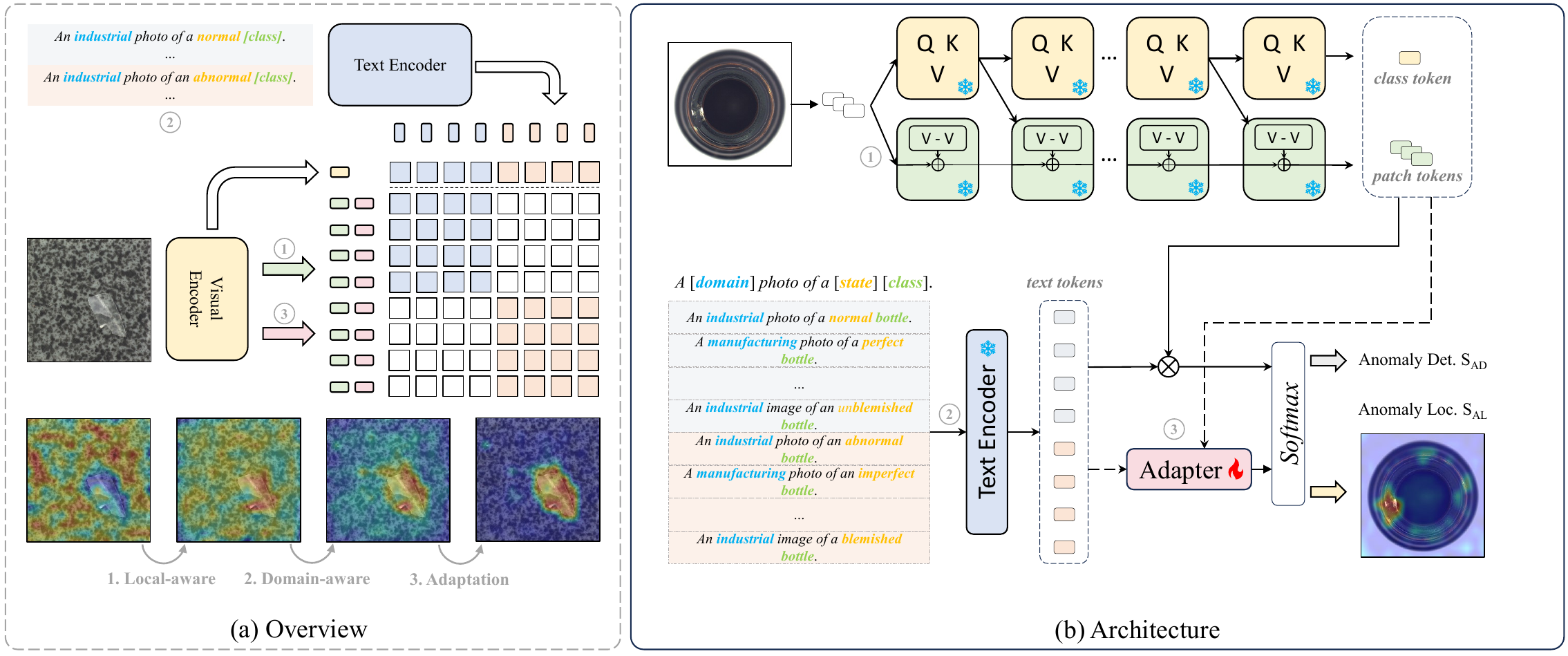} 
\caption{(a) Overview of zero-shot anomaly localization. (b) Our Architecture detail. The solid arrow indicates our AnoCLIP and the dash arrow indicates the procedure of AnoCLIP with TTA. The snow denotes the frozen module and the flame denotes the optimized module. }
\label{framework}
\end{figure*}

To tackle such issues, we propose a novel framework AnoCLIP in Fig. \ref{framework} for adapting CLIP \cite{clip} to zero-shot anomaly localization. 
On one hand, we are concerned with constructing local representations from the pretrained visual model of CLIP. Prior studies show that image-level representations usually originate from the global average pooling operation and CLIP shows proper dense prediction via eliminating attention pooling \cite{cam, maskclip, surgery}. Inspired by this insight, we extract values from the visual encoder and propose to utilize a training-free value-to-value (V-V) attention to compute the local-aware patch tokens. On the other hand, as the zero-shot recognition capability of CLIP depends on the quality of textual prompts \cite{clip, winclip}, we hypothesize that fine-grained detection can benefit from a more precise prompt design and propose a unified domain-aware contrastive state prompt template, ``\emph{A} \ [\textbf{domain}] \ \emph{photo \
 of \ a} \ [\textbf{state}] \ 
 [\textbf{class}]'',
for generating exhaustive prompts. The contrastive state words enable the model to robustly distinguish between normal and abnormal features, and domain-aware prompts encourage textual tokens to adapt to the visual feature space. 
To adapt the visual encoder for downstream domains, we propose a novel fast test-time adaptation (TTA) mechanism for AnoCLIP. Our TTA incorporates a learnable residual-like adapter to refine viusal tokens obtained from AnoCLIP for specific queries. By jointly optimized with two discriminative tasks on AnoCLIP's pseudo-labels and synthetic noise-corrupted tokens, the proposed adapter greatly improves the performance of zero-shot anomaly localization and leverages a proper trade-off between performance and cost. Noticeably, our approach motivates CLIP, a vision-language model for zero-shot image classification, into AnoCLIP for tackling fine-grained anomaly localization. In summary, our contributions are listed as follows:
\begin{enumerate}

\item Compared to one-class and multi-class anomaly detection, we aim at the more scalable zero-shot anomaly detection task, and we proposed a novel unified framework AnoCLIP to tackle this issue. 

\item We address the shortcomings of CLIP in terms of fine-grained local recognizability. With local-aware visual tokens, domain-aware prompting, and a specific test-time adaptation method, our AnoCLIP achieves the ability to localize visual anomalies in the open world.

\item Extensive experiments on MVTecAD \cite{mvtec} and VisA \cite{visa} demonstrate the superiority of our method under the zero-shot anomaly detection and localization setting. We also conduct comprehensive ablation studies to demonstrate the effectiveness of our framework.
\end{enumerate}
\section{Related Work}
\paragraph{Vision-Language Models.} 
As data scale increases, pre-trained visual language models present significant achievements in downstream tasks \cite{vl1,vl2,vl3,clip}. The CLIP models \cite{clip}, derived from contrastive vision language pre-training on millions of image-text pairs gathered from the internet, showcases a profound generality, whereby its prompt-driven zero-shot inference exhibits superior ability for identifying unseen images \cite{cocoop}. Recent studies expand the zero-shot transfer-ability of CLIP models to open-vocabulary semantic segmentation by extracting the intrinsic dense features \cite{maskclip, shin2022reco, surgery}. Some efforts have made considerable progress in improving the recognition performance of CLIP, such as the prompt engineering \cite{coop,cocoop,shu2022test}, the adapter module \cite{adapter, zhang2021tip}, etc. Notably, CLIP can naturally detect out-of-distribution data without any further training, which inspired us to use CLIP for zero-shot anomaly detection and localization.
\paragraph{Anomaly Detection \& Localization}
Most existing anomaly detection methods rely on modeling the distribution of normal samples to identify anomalies \cite{defard2021padim, core, rd, uniad}, whereby prevalent anomaly detection benchmarks typically provide normal-only datasets for training \cite{mvtec, visa}. In particular, robust anomaly detection models often tend to rely on visual models \cite{he2016deep, zagoruyko2016wide, dosovitskiy2020image} pre-trained on large-scale image datasets \cite{deng2009imagenet}. These methods locate anomalies by comparing samples with normal features obtained by reconstruction \cite{mkd, rd, uniad}, caching \cite{core, defard2021padim}, etc. A further challenge is the unified anomaly detection model, which concerns using one model to detect anomalies in multi-class images \cite{regad, uniad}. For the zero-shot setting, some methods attempt to detect internal anomalies using the natural structure of the image \cite{aota2023zero, schwartz2022maeday}. Recently, WinCLIP \cite{winclip} is firstly proposed to process images by multi-scale window moving and classifying each window with CLIP. However, WinCLIP is notably time-consuming due to the additional process of multi-scale windowed image patches. In contrast, our framework leverages CLIP to extract intrinsic dense features directly, enabling more efficient zero-shot anomaly localization.

\section{Methodology}
In this section, we first introduce the CLIP-based baseline model for zero-shot anomaly detection and localization. Then, we elaborate the proposed training-free adaptation, which includes the V-V attention for local-aware token calculation and a domain-aware state prompt template for better vision-language alignment. Lastly, we specify our test-time adaptation mechanism to enhance anomaly localization. Fig. \ref{framework} depicts the overview of our framework for zero-shot anomaly localization.

\subsection{CLIP for Zero-shot Anomaly Recognition}
Let's first briefly review the zero-shot visual recognition ability of CLIP. CLIP includes a visual encoder and a text encoder to extract visual features and text features respectively. Given a comprehensive multi-million dataset of image-text pairs, CLIP trains both encoders by maximizing the feature cosine similarity between images and their corresponding text, which empowers CLIP to align an image with any text in the open vocabulary. For any class in testing, we define a text that includes the concept of this target, e.g. ``a photo of a [class]". Specifically, given an unknown image and a predefined text, the visual encoder and the text encoder output the visual class token $v\in R^C$ and the text token $t\in R^C$, respectively. Here, $C$ repreents the feature dimension. The cosine distance between $v$ and $t$, denoted by $<v, t>$, quantifies the similarity between the image and the class concept.

Anomaly detection involves the semantic concepts of ``normal" and `` anomaly", so for a test class we can simply define two textual prompts i.e. ``a photo of a normal [class]" and ``a photo of an abnormal [class]", and extract the corresponding text tokens [$t^+$, $t^-$]. Given a test image and the corresponding visual token, the probability for anomaly detection is computed as
\begin{equation}
    S_{AD}(v) = \frac{exp(<v,t^->)}{exp(<v,t^+>)+exp(<v,t^->)}.
\end{equation}
Note that no knowledge of visual anomaly is injected into the model, but rather unknown anomalies are detected through the powerful open-world generalization of CLIP.

Although CLIP is only trained to match the global content of an image with text, the final layer of the visual encoder has a set of patch tokens $P=\{p_1,...,p_M|p_i\in R^C\}$ that potentially contain image local information in the patch level. For a patch token $p_i$, the local anomaly score is computed as:
\begin{equation}
    S_{AL}(p_i) = \frac{exp(<p_i,t^->)}{exp(<p_i,t^+>)+exp(<p_i,t^->)}.
\label{eqn:2}
\end{equation}
For an image with $M$ patches split by the visual encoder, the $M$ local anomaly scores are reorganized and up-sampled to align with the original size of the image for a final abnormality map. Note, in such a trivial anomaly localization strategy as a baseline, a lack of alignment between local patch tokens and text embedding leads to limited performance on locating anomalous regions.

\subsection{Extract Local-aware Visual Tokens from CLIP}
The Q-K-V attention mechanism of the visual transformer evenly predisposes patch tokens to characterize global features, since CLIP learns from image-text comparisons at the global level. To extract informative local features from the visual transformer, we carefully modify the attention so as to obtain local patch tokens without any further training. The original Q-K-V attention recipe consists of a layer norm (LN), a Q-K-V projection layer (QKVProj.), an attention mechanism with a projection layer (Proj.), and a multi-layer perceptron (MLP). It can be formulated as:  
\begin{equation}
    Z^{l-1} = [v; p_1; ...; p_M],
\label{eqn:3}
\end{equation}
\begin{equation}
    [Q^l, K^l, V^l] = QKVProj.(LN(Z^{l-1})),
\label{eqn:4}
\end{equation}
\begin{equation}
    Z^{'l} = Proj.(Attention(Q^l, K^l, V^l))+Z^{l-1},
\label{eqn:5}
\end{equation}
\begin{equation}
Z^l = MLP(Z^{'l})+Z^{'l}.
\label{eqn:6}
\end{equation}
Queries and keys in the attention mechanism play the role of associating contextual relations. 
Prior studies show that the Q-K retrieval mode in the last layer somehow acts as a global average pooling for visual global description \cite{maskclip,surgery}. Through bypassing Q-K attention, the values of the last layer lead to local tokens strongly associated with the given text (which has been confirmed by our ablation in Tab. \ref{tab:arch}). To further improve specific locality in patch tokens, we propose to use the value-to-value (V-V) self-attention followed by residual linking per layer to enhance local tokens. Our V-V attention in the adapted $l$th layer of a visual transformer can be formulated as:
\begin{equation}
\tilde{V}^{l} = Proj.(Attention(V^{l-1}, V^{l-1}, V^{l-1}))+\tilde{V}^{l-1},
\end{equation}
where $\tilde{V}^{l}$ denotes the local-aware path of visual encoder. Compared to the original Q-K-V attention in \cref{eqn:3,eqn:4,eqn:5,eqn:6}, V-V attention replaces queries and keys by values and removes the top MLP layer. The patch tokens taken from the last layer of the local-aware path are fed into \cref{eqn:2} for local anomaly scores. Notably, the V-V attention is training-free and runs efficiently for zero-shot anomaly localization.

\subsection{Domain-aware State Prompting}
In contrast to traditional visual models, the predictions of visual language models are critically dominated by the semantic texts. Proper semantic prompts enable CLIP to adapt to a specific task setting with no further training. Especially, anomaly detection, a fine-grained visual recognition task, could benefit from more precise specific prompts. Artificially designing a perfect task-specific prompt sentence is nearly impossible or requires extensive effort, yet a well-designed template can automatically generate enough prompts to comprehensively cover task-related concepts. To this end, we decompose the prompting engineering for a given \textbf{class} of zero-shot anomaly localization into three parts: \textbf{base} prompts, \textbf{contrastive-state} prompts, and \textbf{domain-aware} prompts.

\textbf{Base} prompts are generated from the default prompt ``a photo of a [class]" by varying the context, e.g. ``a cropped photo of a [class]" and ``a bright photo of a [class]". Experiments on CLIP show that the ensemble of multiple base templates boosts the performance of zero-shot classifier.

\textbf{Contrastive-state} (CS) prompts emphasize the antagonistic concepts of normal and abnormal states. According to $S_{AD}$ and $S_{AL}$, we compute a normalized anomaly score with dual text tokens. Thus, we select a series of common opposing state words, such as ``perfect" vs. ``imperfect" and ``with flaw" vs. ``flawless". Especially, anomalies are assumed to be unknown, so we tend to choose common state words to ensure the conceptual coherence and unity. 

\textbf{Domain-aware} (DA) prompting is proposed for bridging the domain gap between the CLIP and the downstream task. Images for fine-grained vision tasks are typically specific distributions, such as industrial images used for visual inspection. To align the distribution of text tokens with visual tokens, we propose domain-aware prompt engineering to adapt prompts to specific domains. For example, ``industrial photo" for industry products, and ``textual photo" for textured structures. As opposed to domain-agnostic prompts, domain-aware prompt engineering eliminates the distribution shift with a non-parametric manner.

Eventually, we summarize the elements that are beneficial to the prompting engineering into a unified template:
$$
\textcolor{gray}{\textbf{A}} \ [\textcolor{Cyan}{\textbf{domain}}] \ \textcolor{gray}{\textbf{photo \
 of \ a}} \ [\textcolor{Dandelion}{\textbf{state}}] \ 
 [\textcolor{YellowGreen}{\textbf{class}}].
$$
Based on this unified prompt template, we provide a few base prompt words and contrast state words and domain-aware words to generate massive prompters for zero-shot anomaly detection and localization. For normal text tokens $[t^+_1, t^+_2, ..., t^+_N]$ and abnormal text tokens $[t^-_1, t^-_2, ..., t^-_N]$ extracted from the prompts list, we compute the averaged tokens as the zero-shot anomaly detector in $S_{AD}$ and $S_{AL}$: 
\begin{equation}
t^+ = \frac{1}{N}\sum^{N}_{i} t^+_i, \ t^- = \frac{1}{N}\sum^{N}_{i} t^-_i.
\end{equation}
We show in experiments that such domain-aware state tokens $[t_+, t_-]$ reliably improve the performance of zero-shot anomaly detection and localization.

\begin{figure}[t]
\centering
\includegraphics[width=\columnwidth]{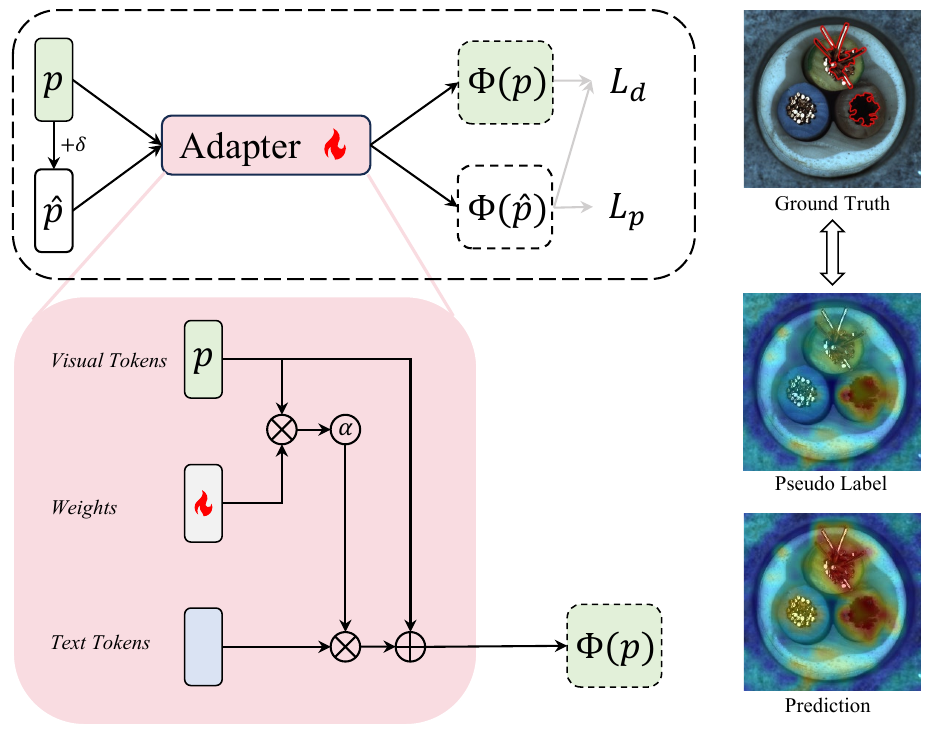} 
\caption{Detailed structure of our adapter. \textit{Weights} denote the learnable paramters of the adapter. We optimize the discriminative objective $L_d$ with adapted patch tokens $\Phi(p)$ and $\Phi(\hat{p})$ and jointly perform pseudo-supervised optimization $L_p$ with $\Phi(p)$.}
\label{adapter}
\end{figure}

\begin{figure*}[ht]
\centering
\includegraphics[width=0.95\textwidth]{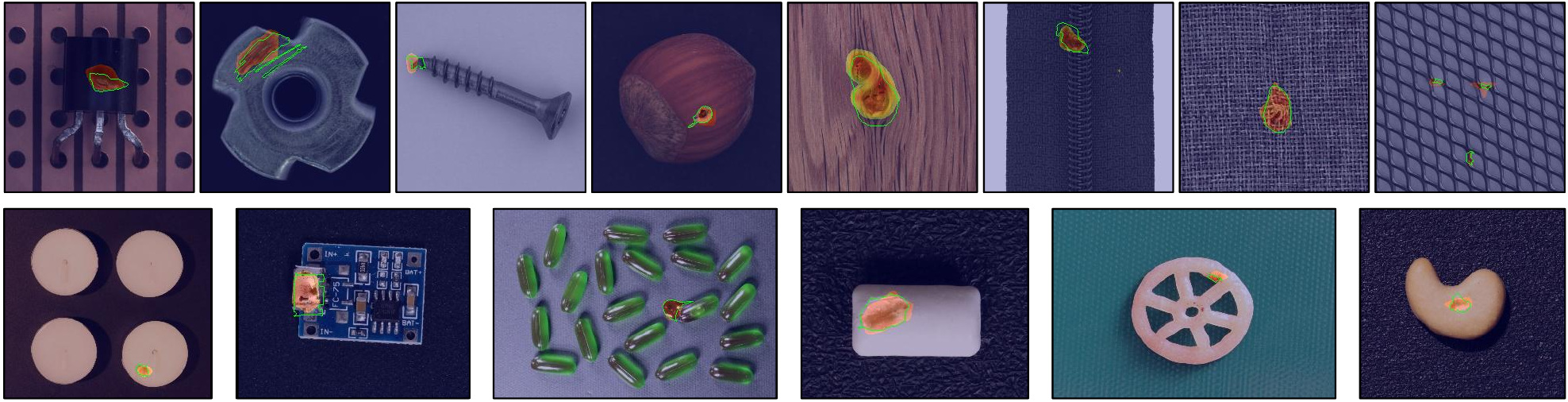} 
\caption{Visualization of zero-shot anomaly localization. The top row shows images from MVTec \cite{mvtec} and the bottom row shows images from VisA \cite{visa}. The red mask refers to our prediction results while the green contour is the ground truth. }
\label{visual}
\end{figure*}

\subsection{Test-time Adaptation for Anomaly Localization}
In addition to eliminating the domain shift of text embedding, we also consider the impact of CLIP's visual distribution shift on downstream tasks. Thus, we aim to use test-time adaptation (TTA) to adapt the visual tokens. TTA is a technique adapting pre-trained models to downstream tasks in inference by optimizing the model with different data views\cite{tta1,tta2,tta3}. However, we identify that previous methods for fine-tuning CLIP on anomaly detection suffer from the following three limitations: (1) End-to-end optimization of all parameters in CLIP typically leads to model collapse \cite {wortsman2022robust}. (2) TTA generally relies on data augmentation to build auxiliary tasks, whereas anomaly detection requires special proxy tasks \cite{cutpaste,draem}. (3) Most of the existent methods for adapting anomaly detection models \cite{cutpaste,draem} and CLIP models \cite{shu2022test} are time-consuming and unfavorable for implementation. To address such issues, we firstly propose a fast TTA strategy for anomaly detection tasks.

Current studies in fine-tuning CLIP focus on text prompt optimization \cite{coop, cocoop} and the visual adapter \cite{adapter}, where prompt optimization takes significantly more time than the adapter. Given that we enable domain adaptation via prompt engineering, and the importance of real-time optimization for TTA, an adapter-like module with few learnable parameters is more adequate for our task. To further tailor the patch tokens for visual-language alignment in anomaly localization with respect to a specific query image, we design a non-linear residual-like adapter $\Phi(\cdot)$ whose structure is shown in Fig. \ref{adapter}.
Mathematically, we denote the set of text prompt tokens as $T \in R^{2N\times C}$, where $t_i$ and $t_{i+N}$ correspond to paired normal and abnormal tokens for $0<i\leq N$. After obtaining the set of patch tokens $P\in R^{M\times C}$ for a test image, 
the online adaptation can be written as:
\begin{equation}
\Phi(P) = \alpha (P\omega^\mathsf{T})T+P,
\label{eqn:TTA}
\end{equation}
where $\alpha$ is the activation function and $\omega \in R^{2N\times C}$ denotes the learnable parameters initialized by the text tokens $T$. Notably, the proposed adaptor in Eq. (\ref{eqn:TTA}) can be considered an attention operation taking $P, \omega, T$ as Q, K and V, respectively. It enhances the alignment between text prompts and local patches by injecting semantic information in $P'$.

Optimizing the trainable parameter $\omega$ in the adaptor involves in designing self-supervised tasks. In contrast to traditional TTAs that use standard data augmentation \cite{gidaris2018unsupervised, hendrycks2019augmix,devries2017improved}, anomaly detection tasks typically take the form of image partial corruption \cite{cutpaste,draem}. However, image-level augmentation still requires multiple-pass of forward propagation, which is time-consuming for real-time anomaly localization. By considering that feature-level augmentation only involves passing through a lightweight adapter, so we propose to directly augment visual tokens by perturbation. In this study, We use two self-supervised discriminative tasks are constructed to optimize $\omega$. Firstly, learning to discriminate between given and forged features facilitates the adapter in distinguishing anomalous representations \cite{simple}. Specifically, for patch tokens $P=\{p_1,..., p_M\}$ from a query, we synthesize noise-corrupted tokens $\hat{P}=\{\hat{p}_1,..., \hat{p}_M\}$, where $\hat{p}_i = p_i + \delta_i$ and $\delta_i \in R^C$ is Gaussian noise sampled from $\mathcal{N}(\mu, {\sigma}^2)$. 
To discriminate original and noise-corrupted tokens in $P$ and $\hat{P}$, a discriminitive loss over the $2M$ patch tokens: 
\begin{equation}
    L_d = -\frac{1}{2M}\sum^{M}_{i=1}[\mathsf{log}(1-S_{AL}(\Phi(p_i)))+\mathsf{log}S_{AL}(\Phi(\hat{p}_i))],
\label{eqn:10}
\end{equation}
encourages the adaptor being sensitive to subtle changes on patch tokens. 
However, \cref{eqn:10} only distinguishes between true and noisy distributions, and we still need to get the adapter to learn to distinguish between true normal and abnormal feature distributions for keeping balance. To encourage the adapter to preserve real normal and abnormal features, we take the anomaly probability map predicted by the AnoCLIP on $P$ as pseudo-labels. Then, we define a soft cross-entropy loss as below:
\begin{equation}
    L_p = -\frac{1}{M}\sum^{M}_{i=1}S_{AL}(\Phi(p_i))\mathsf{log}(S_{AL}(\Phi(p_i))).
\end{equation}
Inspired by self-distillation \cite{hamilton2022unsupervised}, pseudo-label learning can generate more compact downstream features. The two optimization objectives enable a balance of test-time refinement for anomaly detection, which is to achieve visual domain adaptation as well as to enhance anomaly discrimination.

The overall learning objective to update $\omega$ in our adaptor is $L = L_d+L_p$. Our TTA strategy is time-efficient and  can significantly improve the zero-shot anomaly localaztion performance of AnoCLIP without any training data and annotation. Note that we use AnoCLIP+ to denote AnoCLIP optimized by TTA in the experiment section. 

\section{Experiment}
\subsection{Experimental Setup} 
\paragraph{Datasets.} We conduct zero-shot anomaly detection and localization experiments on two datasets: MVTecAD \cite{mvtec} and VisA \cite{visa}. MVTecAD consists of data for 10 single objects and 5 textures, while VisA includes data for 12 single or multiple object types. In this paper, only the test datasets are used for evaluating zero-shot anomaly detection and localization, and no extra datasets are available.
\paragraph{Metrics.}Following conventions in prior arts \cite{mvtec, winclip}, we use AUROC (Area Under the Receiver Operating Characteristic), F1Max, and PRO (Per-Region Overlap) to evaluate anomaly detection. Besides, we use AUROC, F1Max, and AUPR ( Area Under the Precision-Recall curve) to evaluate anomaly localization.
\paragraph{Implementation.} 
We adopt ViT-B-16+ \cite{dosovitskiy2020image} as the visual encoder and the transformer \cite{vaswani2017attention} as the text encoder by default from the public pre-trained CLIP model \cite{ilharco_gabriel_2021_5143773}. We also provide results under ResNet \cite{he2016deep} and other ViT \cite{dosovitskiy2020image} models \cite{clip, ilharco_gabriel_2021_5143773} in the supplement. For the visual encoder, we set the image size to $240\times 240$ without any image augmentation. For the text encoder, we use $22$ base templates collected from CLIP \cite{clip}, $7$ pairs of state prompts, and $4$ domain-aware prompts to generate sufficient prompts on the proposed template of domain-aware state prompting. All prompts are listed in the supplement. For TTA, we use the AdamW optimizer and set the learning rate to $0.001$, and the adaptor is optimized with $5$ epochs. We set $\mu=0$ and $\sigma=0.05$ for the i.i.d Gaussion noise $\mathcal{N}(\mu, {\sigma}^2)$ used in TTA. Since AnoCLIP+ involves optimizing AnoCLIP via TTA, We report the mean and variance of the results of AnoCLIP+ over $5$ random seeds.

\begin{table}[ht]
\centering
\fontsize{7.5}{15}\selectfont
\setlength\tabcolsep{1pt}
\begin{tabular}{lcccccc}
\hline
AL         & \multicolumn{3}{c}{MVTecAD \cite{mvtec}}                               & \multicolumn{3}{c}{VisA \cite{visa}}                                  \\ \hline
Method     & AUROC             & F1Max             & PRO               & AUROC             & F1Max             & PRO               \\
\cmidrule(lr){1-1}\cmidrule(lr){2-4}\cmidrule(lr){5-7}
TransMM\cite{transmm}    & 57.5              & 12.1              & 21.9              & 49.4              & 14.8              & 10.2              \\
MaskCLIP\cite{maskclip}   & 63.7              & 18.5              & 40.5              & 60.9              & 7.3               & 27.3              \\
WinCLIP\cite{winclip}    & 85.1              & \underline{31.7}        & 64.6              & 79.6              & \underline{14.8}        & 59.8              \\
CLIP\cite{clip}       & 19.5              & 6.2               & 1.6               & 22.3              & 1.4               & 0.8               \\
AnoCLIP & \underline{86.6}        & 30.1              & \underline{70.4}     & \underline{85.9}     & 14.7              & \underline{65.1}     \\
AnoCLIP+ & \textbf{90.6$\pm$0.3} & \textbf{36.5$\pm$0.3} & \textbf{77.8$\pm$0.2} & \textbf{91.4$\pm$0.2} & \textbf{17.4$\pm$0.4} & \textbf{75.0$\pm$0.1} \\ \hline
\end{tabular}
\caption{Zero-shot anomaly localization (AL) on MVTecAD and VisA. Bold indicates the best performance and underline indicates the runner-up unless other noted.}
\label{tab:al}
\end{table}

\begin{table}[ht]
\centering
\fontsize{8}{15}\selectfont
\setlength\tabcolsep{3pt}
\begin{tabular}{lcccccc}
\hline
AD & \multicolumn{3}{c}{MVTecAD \cite{mvtec}}                     & \multicolumn{3}{c}{VisA \cite{visa}}                      \\ \hline
Method            & AUROC         & F1Max         & AUPR          & AUROC         & F1Max         & AUPR          \\ \cmidrule(lr){1-1}\cmidrule(lr){2-4}\cmidrule(lr){5-7}
CLIP\cite{clip}              & 74.0          & 88.5          & 89.1          & 59.3          & 74.4          & 67.0          \\
WinCLIP\cite{winclip}           & \underline{91.8}          & \underline{92.9}          & \underline{96.5}          & \underline{78.1}          & \underline{79.0}          & \underline{81.2}          \\
AnoCLIP              & \textbf{92.5} & \textbf{93.2} & \textbf{96.7} & \textbf{79.2} & \textbf{79.7} & \textbf{81.7} \\ \hline
\end{tabular}
\caption{Zero-shot anomaly detection (AD) performance. Note that AnoCLIP+ has the same anomaly detection performance as AnoCLIP, since TTA only refines the localization.}
\label{tab:ad}
\end{table}

\subsection{Performance}
Table \ref{tab:al} and Table \ref{tab:ad} present the performance of zero-shot anomaly localization and detection on MVTecAD and VisA, respectively. We compare our proposed methods with prior vision-language based works, including vanilla CLIP \cite{clip}, TransMM \cite{transmm}, MaskCLIP \cite{maskclip}, and WinCLIP \cite{winclip}. Remarkablely, both our proposed AnoCLIP and AnoCLIP+ show outstanding performance over other methods, 
significantly surpassing WinCLIP by margins of $5.5\%$, $4.8\%$ and $13.2\%$ in terms of AUROC, F1Max, and PRO, respectively. 
In Fig. \ref{visual}, we demonstrate the results of zero-shot anomaly localization. Both the visualizations and experimental results illustrate that our unified zero-shot anomaly localization model possesses the ability to effectively localize various types of defects across diverse samples.

\subsection{Ablation Studies}
\paragraph{Local-aware tokens extraction.} In Table \ref{tab:arch}, we present an analysis of the impact derived from modifying the vision attention within CLIP, a pivotal step towards acquiring local-aware tokens for anomaly localization. In comparison to the QKV attention mechanism in CLIP, the values (V) extracted from the final layer of the transformer as patch tokens shows basic anomaly localization capabilities. Notably, the patch tokens acquired through the residual accumulation of V-V attention across the multiple layers yield promising anomaly localization performance.
\begin{table}[ht]
\centering
\fontsize{7}{15}\selectfont
\setlength\tabcolsep{10pt}
\begin{tabular}{lccc}
\cline{1-4}
MVTecAD \cite{mvtec}      & AUROC & F1Max & PRO \\ \cmidrule(lr){1-1}\cmidrule(lr){2-4}
with QKV attention &  19.5     &   6.2    &   1.6    \\
with V (last layer)    &   67.9    &  22.2     &  50.6  \\
with V-V attention (last layer)  &  70.5     &  22.0     &     51.9     \\
with V-V attention (multi-layer) &  \textbf{86.6}     &  \textbf{30.1}     & \textbf{70.4}      \\ \cline{1-4}
\end{tabular}
\caption{Ablation studies for the attention mode of CLIP.}
\label{tab:arch}
\end{table}
\paragraph{Prompt engineering.} Given the significance of language prompts in steering CLIP toward effective anomaly detection and localization, we undertake an analysis to the prompt engineering. The results of this examination are elaborated in Table \ref{tab:prompt}. 
By designing multiple contrastive state prompts, we obtain a dramatic gain in both anomaly detection and localization. 
Besides, domain-aware prompting enables the direct adaptation of text tokens to diverse data distributions, making a substantial contribution to the improvement in performance.

\begin{figure}[ht]
\centering
\includegraphics[width=\columnwidth]{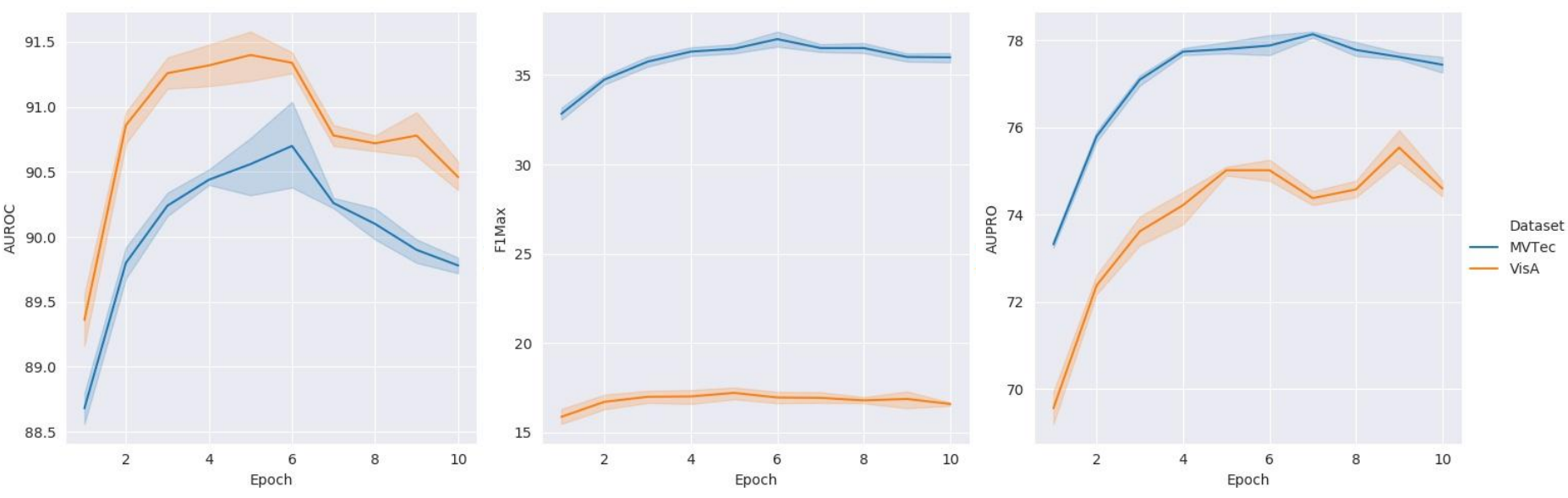} 
\caption{Ablation on TTA epochs on MVTecAD and VisA. We measure AUROC, F1Max, and PRO results for AnoCLIP+ within 10 epochs on MVTecAD and VisA.}
\label{epoch}
\end{figure}

\begin{table}[ht]
\centering
\fontsize{8}{15}\selectfont
\setlength\tabcolsep{4pt}
\begin{tabular*}{\columnwidth}{lllllll}
\hline
MVTecAD \cite{mvtec}   & \multicolumn{3}{c}{Zero-shot A.D.} & \multicolumn{3}{c}{Zero-shot A.L.} \\ \hline
Method        & AUROC     & F1Max     & AUPR     & AUROC     & F1Max     & PRO    \\ \cmidrule(lr){1-1} \cmidrule(lr){2-4}\cmidrule(lr){5-7}
Base         &   79.0         & 89.7          &  91.7        &  71.9        & 27.0         &    50.5      \\
+CS Prompt    &  91.8         &  92.8         &   96.4       &   84.2        &   28.7        &  67.4        \\
+DA Prompt &      \textbf{92.5}     &    \textbf{93.2}       &    \textbf{96.7}      &     \textbf{86.6}      &     \textbf{30.1}      &    \textbf{70.4}      \\ \hline
\end{tabular*}
\caption{Ablation studies for prompt engineering. Note that our prompt engineering is also effective for anomaly detection (AD).}
\label{tab:prompt}
\end{table}

\begin{table*}[!t]
\centering
\fontsize{8}{13}\selectfont
\setlength\tabcolsep{5pt}
\begin{tabular}{lcccccccccc}
\hline
Setting & \multicolumn{8}{c}{Full-shot}                                             & \multicolumn{2}{c}{Zero-shot} \\ \cmidrule(lr){1-1}\cmidrule(lr){2-9}\cmidrule(lr){10-11}
Method  & US\cite{us}   & PSVDD\cite{psvdd} & PaDiM\cite{defard2021padim} & CutPaste\cite{cutpaste} & FCDD\cite{fcdd} & MKD\cite{mkd}  & DRAEM\cite{draem} & UniAD\cite{uniad}             & WinCLIP\cite{winclip}    & AnoCLIP+             \\ 
AD      & 74.5 & 76.8  & 84.2  & 77.5     & -    & 81.9 & 88.1  & \textbf{96.5+0.1} & 91.2       & \underline{92.5+0.0}   \\
AL      & 81.8 & 85.6  & 89.5  & -        & 63.3 & 84.9 & 87.2  & \textbf{96.8+0.0} & 85.1       & \underline{90.6+0.3}   \\ \hline
\end{tabular}
\caption{Comparision of multi-class anomaly detection and localization performance on MVTecAD. We compare the performance of our zero-shot approach to the state-of-the-art full-shot unified model UniAD \cite{uniad}, as well as US \cite{us}, PSVDD \cite{psvdd}, PaDiM \cite{defard2021padim}, CutPaste \cite{cutpaste}, FCDD \cite{fcdd}, MKD \cite{mkd}, and DRAEM \cite{draem} on the multi-class task. Full-shot results are taken from UniAD \cite{uniad}.}
\label{tab:lim}
\end{table*}

\begin{table}[ht]
\centering
\fontsize{8}{15}\selectfont
\setlength\tabcolsep{12pt}
\begin{tabular}{lccc}
\cline{1-4}
MVTecAD \cite{mvtec} & AUROC & F1Max & PRO \\ \cmidrule(lr){1-1}\cmidrule(lr){2-4}
Linear Probe &   88.9$\pm$0.1    & 33.1$\pm$0.2      &  74.4$\pm$ 0.1   \\
Ours    &  \textbf{90.6$\pm$0.3}     &   \textbf{36.5$\pm$0.3}    &  \textbf{77.8$\pm$0.2}     \\ \cline{1-4}
\end{tabular}
\caption{Ablation studies on TTA adapter optimization.}
\label{tab:adapter}
\end{table}

\begin{figure}[ht]
\centering
\includegraphics[width=0.9\columnwidth]{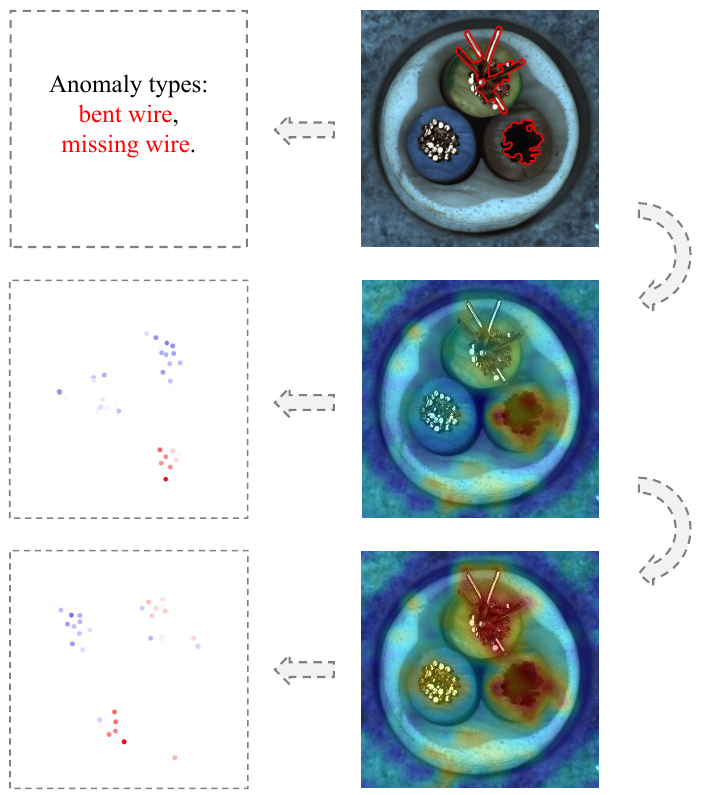} 
\caption{t-SNE visulization on AnoCLIP and AnoCLIP+. The sample is from $``cable"$ in MVTecAD with anomaly types: bent wire and missing wire. Red indicates abnormal state and blue indicates normal state, while the shade represent the anomaly score. }
\label{tfa_tta}
\end{figure}

\paragraph{Test-time adaptation with dual-task optimization.} The CLIP-Adapter \cite{adapter} proposed to fine-tune a linear probe on few samples for image recognition. Nevertheless, our test-time adaptation exhibits only marginal improvement when employed with the linear probe for $\omega$ optimization, due to poor random initialization. 
Therefore, we propose to use the text tokens to align with the weights in this study. 
Table \ref{tab:adapter} shows that our adapter optimized by the proposed dual tasks brings a great performance improvement compared to the linear probe. Additionally, we visualize the impact of running epochs on performance in Fig. \ref{epoch}. As the number of running epochs increases, our TTA stategy shows a trend of convergence. While a slight performance decrease is observed with more iterations, it's important to note that TTA often involves a limited number of epochs due to time constraints. Considering the trade-off between performance and time cost, we exclusively run 5 epochs for TTA within our experiments.
We show the patch tokens virtualisation of AnoCLIP and AnoCLIP+ (with TTA) processed by t-SNE \cite{van2008visualizing} in Fig. \ref{tfa_tta}. In this case, AnoCLIP shows no tendency to some anomalous tokens, while TTA refines them to yield more precise results.

\paragraph{Model Efficiency.}
In Table \ref{tab:time}, we present the zero-shot anomaly localization efficiency concerning performance and inference time. Our AnoCLIP with only 1.4 ms increase in inference time exceeds the performance of WinCLIP, which is time-consuming because of the multi-scale window moving operation. AnoCLIP+ yields a significant performance boost at the cost of $3\times$ time consumption compared to AnoCLIP, while still more efficient than WinCLIP. 
\begin{table}[ht]
\centering
\fontsize{8}{15}\selectfont
\setlength\tabcolsep{6pt}
\begin{tabular}{lcccc}
\cline{1-5}
Methods & AUROC & F1Max & PRO & Time (ms) \\ \cmidrule(lr){1-1}\cmidrule(lr){2-4}\cmidrule(lr){5-5}
WinCLIP\cite{winclip}         &  85.1     &   \underline{31.7}    &  79.6     &  41.3$\pm$1.2         \\
CLIP\cite{clip}            &  19.5     & 6.2      & 1.6      &     \textbf{6.2$\pm$0.2}      \\
AnoCLIP      &  \underline{86.6}     & 30.1      & \underline{70.4}      &  \underline{7.6$\pm$0.2}         \\
AnoCLIP+      &  \textbf{90.6$\pm$0.3}     &  \textbf{36.5$\pm$0.3}     &  \textbf{77.8$\pm$0.2}     &  20.1$\pm$1.4         \\ \cline{1-5}
\end{tabular}
\caption{Comparision of model efficiency on MVTecAD. We measure the inference time per image with an RTX3090 GPU.}
\label{tab:time}
\end{table}

\section{Limitations and Conclusion}

\paragraph{Limitations \& future work}
While we achieve truly unified anomaly detection and localization and SOTA performance in the zero-shot setting, there is still a gap compared to the SOTA model trained in the full-shot manner. Table \ref{tab:lim} shows our approach surpasses most of the models on unified anomaly detection and localization task. However, we underperform the UniAD \cite{uniad}, which achieves the best results by training on normal datasets. Therefore, our future work is to fine-tune visual language models on some normal data, such as few-shot and full-shot, to achieve better performance. Compared to some recent studies on zero-shot anomaly detection via supervised fine-tuning \cite{chen2023clip, chen2023zero} and unsupervised prompt optimization \cite{gu2023anomalygpt}, our approach does not require any training data. In addition, our study brings new insights for fine-tuning CLIP in fine-grained anomaly detection: (1) visual and textual domain adaptation contributes to fine-grained anomaly localization for CLIP. (2) Our decomposition of prompts reveals the focus for textual optimization \cite{coop,cocoop} of visual anomaly detection.

\paragraph{Conclusion}
We proposed AnoCLIP aimed at resolving unified zero-shot anomaly detection and localization. Through adapting the vision-language models such as CLIP, we achieve SOTA performance without any training dataset. We first extract implicitly local-aware tokens from CLIP for precise anomaly localization. Then, we propose a domain-aware state prompting template for generating sufficient and effective prompts. Finally, we use test-time adaptation to refine features to localize anomalies more precisely. This study demonstrates that vision-language models are able to solve unified anomaly detection and localization problems without being constrained by training data.

{
    \small
    \bibliographystyle{ieeenat_fullname}
    \bibliography{main}
}

\end{document}


\maketitle

\section{Dataset}
Our experiments are based on MVTecAD \cite{mvtec} and VisA \cite{visa} datasets. Both datasets contain diverse subsets of different objects and texture categories. Each dataset contains high-resolution image, with dimension ranging from $(700\times700$ to $1024\times1024$ for MVTecAD, whereas VisA employs rectangular images with approximately $1500\times1000$ for resolution, accompanied by corresponding anomaly ground truth masks.

MVTecAD serves as a standard benchmark which contains 15 sub-datasets with a total of 5354 images, featuring 5 categories of texture and 10 of objects. Each sub-dataset is divided into anomaly-free training data and test set contains both nominal and anomalous samples, notebly, the test set includes 1725 images.

Our evaluation extends to the recently published challenging VisA, which is twice the size of MVTecAD, boosting of 9621 nominal and 1200 abnormal images, distributed across 12 subsets of different objects. Anomalous images exhibit a variety of imperfections, ranging from surface defects such as scratches, dents, colored spots, or cracks, to structure defects like misplacement or missing parts.

We apply the data preprocessing pipeline outlined in OpenCLIP \cite{ilharco_gabriel_2021_5143773} for both MVTecAD and VisA benchmarks. This involves bilinear resizing of images and  corresponding pixel annotations to the height of $240$. Subsequently, a channel-wise normalization is performed with $mean=(0.48145466, 0.4578275, 0.40821073), std=(0.26862954, 0.26130258, 0.27577711)$ after scaling the grayscale values into $[0,1]$. Given the non-squared images in VisA,  the resized images don't follow exact shape of $(240, 240)$, potentially leading to train-test discrepancy. To address this, we propose image tiling to duplicate 2 images of size $(240, 240)$ by shifting the window on width dimension. After inference, the 2 images are reintegrated into a single image, with the overlapping region computed as the mean of both image results.

\section{Evaluation Metrics}
We assess the efficacy of our model by utilizing the Area Under Receiver Operator Characteristics (AUROC) image-level AUROC is used for anomay detection, and the pixel-level AUROC is measured for evaluating anomaly localication. However, the AUROC does not reflect the localization accuracy well in anomaly localization setups, where only a small fraction of pixels are anomalous, the metric is dominated by a very high number of non-anomalous pixels and is thus kept high despite of false detections. We thus additionally report the F1-max score and Area Under Precisoin-Recall (AUPR) as a balanced calculation of the precision and recall to overcome the class imbalance. In addition to that, we compute the Per-Region-Overlap (PRO) to measure anomaly localization, which weights each connected component within the ground-truth of varying sizes equally, making it more robust than simple pixel measurement.

\section{Additional Results}
We compare the performance of different backbones as visual encoders on CLIP models from OpenCLIP \cite{ilharco_gabriel_2021_5143773} on MVTecAD in Tables \ref{mvtec_auroc}, \ref{mvtec_f1max}, and \ref{mvtec_pr}. In particular, we adapted only the last layer of attention pooling for ResNet \cite{he2016deep}. Note that we report results for CLIP with our AnoCLIP/AnoCLIP+. Our method performs much better on ViT \cite{dosovitskiy2020image} than on ResNet due to the fact that our AnoCLIP only extracts local features from the last layer when acting on ResNet. Remarkably, our proposed test-time adaptation (AnoCLIP+) can further refine the prediction results, thus improving the performance of ResNet-based CLIP for zero-shot anomaly localization.
\begin{table}[!h]
\centering
\fontsize{8}{15}\selectfont
\setlength\tabcolsep{10pt}
\begin{tabular}{lccc}
\hline
\multicolumn{4}{c}{AUROC - MVTecAD}                                                    \\ \hline
Model                        & Size                         & AL        & AD   \\ \cmidrule(lr){1-1}\cmidrule(lr){2-2}\cmidrule(lr){3-4}
RN50                         & 224$\times$224 & 66.3/81.4$\pm$0.2 & 77.4 \\
RN101                        & 224$\times$224 & 60.8/80.9$\pm$0.2 & 75.6 \\
RN50$\times$4  & 288$\times$288 & 66.2/86.4$\pm$0.3 & 80.6 \\
RN50$\times$16 & 384$\times$384 & 58.8/87.8$\pm$0.2 & 75.2 \\
ViT-B/16                     & 224$\times$224 & 83.8/88.8$\pm$0.3 & 87.9 \\
ViT-B/16+                    & 240$\times$240 & 86.6/90.6$\pm$0.3 & 92.5 \\ \hline
\end{tabular}
\caption{Comparison of zero-shot anomaly detection (AD) and anomaly localization (AL) performance in AUROC on   MVTecAD, across various visual encoders of CLIP.}
\label{mvtec_auroc}
\end{table}

\begin{table}[!h]
\centering
\fontsize{8}{15}\selectfont
\setlength\tabcolsep{10pt}
\begin{tabular}{lccc}
\hline
\multicolumn{4}{c}{F1Max - MVTecAD}                                            \\ \hline
Model                        & Size                         & AL        & AD   \\ \cmidrule(lr){1-1}\cmidrule(lr){2-2}\cmidrule(lr){3-4}
RN50                         & 224$\times$224 & 13.7/18.7$\pm$0.4 & 88.1 \\
RN101                        & 224$\times$224 & 15.2/19.2$\pm$0.4 & 87.9 \\
RN50$\times$4  & 288$\times$288 & 20.0/25.1$\pm$0.2 & 89.2 \\
RN50$\times$16 & 384$\times$384 & 18.3/29.5$\pm$0.3 & 86.5 \\
ViT-B/16                     & 224$\times$224 & 27.1/34.2$\pm$0.2 & 90.8 \\
ViT-B/16+                    & 240$\times$240 & 30.1/36.5$\pm$0.3 & 93.2 \\ \hline
\end{tabular}
\caption{Comparison of zero-shot anomaly detection (AD) and anomaly localization (AL) performance in F1Max on   MVTecAD, across various visual encoders of CLIP.}
\label{mvtec_f1max}
\end{table}

\begin{table}[!h]
\centering
\fontsize{8}{15}\selectfont
\setlength\tabcolsep{10pt}
\begin{tabular}{lccc}
\hline
\multicolumn{4}{c}{PRO (AL) / AUPR (AD) - MVTecAD}                             \\ \hline
Model                        & Size                         & AL        & AD   \\ \cmidrule(lr){1-1}\cmidrule(lr){2-2}\cmidrule(lr){3-4}
RN50                         & 224$\times$224 & 36.0/54.7$\pm$0.3 & 89.0 \\
RN101                        & 224$\times$224 & 34.6/50.8$\pm$0.2 & 88.9 \\
RN50$\times$4  & 288$\times$288 & 37.0/62.2$\pm$0.3 & 91.2 \\
RN50$\times$16 & 384$\times$384 & 37.5/69.8$\pm$0.3 & 88.3 \\
ViT-B/16                     & 224$\times$224 & 65.7/75.2$\pm$0.1 & 93.8 \\
ViT-B/16+                    & 240$\times$240 & 70.4/77.8$\pm$0.2 & 96.7 \\ \hline
\end{tabular}
\caption{Comparison of zero-shot anomaly detection (AD) in AUPR and anomaly localization (AL) performance in PRO on MVTecAD, across various visual encoders of CLIP.}
\label{mvtec_pr}
\end{table}

\section{Prompt Engineering}
In this section, we provide details on our choice of base templates, contrastive state words and domain-aware prompts. We denote each class as [c], the state prompts as [s], and the domain-aware prompts as [d]. By replacing [s] and [d] into [c], we obtain lots of normal and anomalous prompts.
\paragraph{Base Templates}
\begin{enumerate}
\item ``a [d] cropped photo of the [s]"\\
\item ``a [d] cropped photo of a [s]"\\
\item ``a [d] close-up photo of a [s]"\\
\item ``a [d] close-up photo of the [s]"\\
\item ``a [d] bright photo of a [s]"\\
\item ``a [d] bright photo of the [s]"\\
\item ``a [d] dark photo of the [s]"\\
\item ``a [d] dark photo of a [s]"\\
\item ``a [d] jpeg corrupted photo of a [s]"\\
\item ``a [d] jpeg corrupted photo of the [s]"\\
\item ``a [d] blurry photo of the [s]"\\
\item ``a [d] blurry photo of a [s]"\\
\item ``a [d] photo of a [s]"\\
\item ``a [d] photo of the [s]"\\
\item ``a [d] photo of a small [s]"\\
\item ``a [d] photo of the small [s]"\\
\item ``a [d] photo of a large [s]"\\
\item ``a [d] photo of the large [s]"\\
\item ``a [d] photo of the [s] for visual inspection"\\
\item ``a [d] photo of a [s] for visual inspection"\\
\item ``a [d] photo of the [s] for anomaly detection"\\
\item ``a [d] photo of a [s] for anomaly detection"\\
\end{enumerate}
\paragraph{Contrastive State Prompts}
Normal states:\\
\begin{enumerate}
\item s := ``normal [c]"\\
\item s := ``unblemished [c]"\\
\item s := ``flawless [c]"\\
\item s := ``perfect [c]"\\
\item s := ``[c] without flaw"\\
\item s := ``[c] without defect"\\
\item s := ``[c] without damage"\\
\end{enumerate}

\begin{figure}[!t]
\includegraphics[width=\columnwidth]{vis 1} 
\caption{Visualization of examples from ``bottle", ``cable", ``capsule", ``hazel nut", and ``metal nut" categories.}
\label{v1}
\end{figure}

Anomaly states:\\
\begin{enumerate}
\item s := ``abnormal [c]"\\
\item s := ``blemished [c]"\\
\item s := ``flawed [c]"\\
\item s := ``imperfect [c]"\\
\item s := ``[c] without flaw"\\
\item s := ``[c] without defect"\\
\item s := ``[c] without damage"\\
\end{enumerate}
\paragraph{Domain-aware Prompts}
\begin{enumerate}
\item d := ``industrial"\\
\item d := ``manufacturing"\\
\item d := ``textural"\\
\item d := ``surface"\\
\end{enumerate}

    
\section{Additional Visualization}
We visualize the zero-shot anomaly localization results on MVTecAD \cite{mvtec} in \cref{v1,v2,v3} and on VisA \cite{visa} in \cref{v4,v5,v6}. In particular, the threshold for segmentation is chosen from the maximum value of the F1 score. Notably, for some weak cases predicted in AnoCLIP, AnoCLIP+ is able to precisely localize the anomalies.

\begin{figure}[!t]
\centering
\includegraphics[width=\columnwidth]{vis 2} 
\caption{Visualization of examples from ``pill", ``screw", ``toothbrush", ``transistor", and ``zipper" categories.}
\label{v2}
\end{figure}

\begin{figure}[!ht]
\includegraphics[width=\columnwidth]{vis 3} 
\caption{Visualization of examples from ``carpet", ``grid", ``leather", ``till", and ``wood" categories.}
\label{v3}
\end{figure}

\begin{figure}[!ht]
\includegraphics[width=\columnwidth]{vis 4} 
\caption{Visualization of examples from ``candle", ``capsules", ``fryum", and ``chewinggum" categories.}
\label{v4}
\end{figure}
\begin{figure}[!ht]
\includegraphics[width=\columnwidth]{vis 5} 
\caption{Visualization of examples from ``macaroni1", ``macaroni2", ``cashew", and ``pipe fryum" categories.}
\label{v5}
\end{figure}
\begin{figure}[!ht]
\includegraphics[width=\columnwidth]{vis 6} 
\caption{Visualization of examples from ``pcb1", ``pcb2", ``pcb3", and ``pcb4" categories.}
\label{v6}
\end{figure}
{
    \small
    \bibliographystyle{ieeenat_fullname}
    \bibliography{main}
}
